%% file: Unified-VCR.tex
\documentclass[lettersize,journal]{IEEEtran}
\usepackage{amsmath,amsfonts}
\usepackage{algorithm}
\usepackage{array}
\usepackage[caption=false,font=normalsize,labelfont=sf,textfont=sf]{subfig}
\usepackage{textcomp}
\usepackage{stfloats}
\usepackage{url}
\usepackage{verbatim}
\usepackage{graphicx}
\usepackage{cite}
\usepackage{algorithmicx}
\usepackage{algpseudocode}
\usepackage{tabularx}
\usepackage{chngpage}
\usepackage{latexsym}
\usepackage{booktabs}
\usepackage{multirow}
\usepackage{multicol}
\usepackage{color}
\usepackage{dsfont}
\usepackage{makecell}

\newcommand{\ie}{\textit{i}.\textit{e}., }
\newcommand{\eg}{\textit{e}.\textit{g}., }

\begin{document}

\title{Joint Answering and Explanation for Visual Commonsense Reasoning}

\author{Zhenyang Li$\dagger$, 
        Yangyang Guo$\dagger$,~\IEEEmembership{Member,~IEEE},
        Kejie Wang,
        Yinwei Wei,~\IEEEmembership{Member, IEEE},
        Liqiang Nie,~\IEEEmembership{Senior Member,~IEEE},
        Mohan Kankanhalli,~\IEEEmembership{Fellow,~IEEE}
\IEEEcompsocitemizethanks{
\IEEEcompsocthanksitem (Corresponding authors: Yangyang Guo and Liqiang Nie.)
\IEEEcompsocthanksitem $\dagger$ Equal contribution.
\IEEEcompsocthanksitem Zhenyang Li and Kejie Wang are with Shandong University, China. E-mail: \{zhenyanglidz, kjwang.henry\}@gmail.com.
\IEEEcompsocthanksitem Yangyang Guo, Yinwei Wei and Mohan Kankanhalli are with National University of Singapore, Singapore. E-mail: \{guoyang.eric, weiyinwei@hotmail.com, mohan@comp.nus.edu.sg.
\IEEEcompsocthanksitem  Liqiang Nie is with Harbin Institute of Technology (Shenzhen), China. E-mail: nieliqiang@gmail.com.
}}

\markboth{IEEE TRANSACTIONS ON IMAGE PROCESSING}%
{Joint Answering and Explanation for Visual Commonsense Reasoning}


\maketitle

\begin{abstract}
Visual Commonsense Reasoning (VCR), deemed as one challenging extension of Visual Question Answering (VQA), endeavors to pursue a higher-level visual comprehension. 
VCR includes two complementary processes: question answering over a given image and rationale inference for answering explanation. 
Over the years, a variety of VCR methods have pushed more advancements on the benchmark dataset. 
Despite significance of these methods, they often treat the two processes in a separate manner and hence decompose VCR into two irrelevant VQA instances.
As a result, the pivotal connection between question answering and rationale inference is broken, rendering existing efforts less faithful to visual reasoning.
To empirically study this issue, we perform some in-depth empirical explorations in terms of both language shortcuts and generalization capability.
Based on our findings, we then propose a plug-and-play knowledge distillation enhanced framework to couple the question answering and rationale inference processes. 
The key contribution lies in the introduction of a new branch, which serves as a relay to bridge the two processes. 
Given that our framework is model-agnostic, we apply it to the existing popular baselines and validate its effectiveness on the benchmark dataset. 
As demonstrated in the experimental results, when equipped with our method, these baselines all achieve consistent and significant performance improvements, evidently verifying the viability of processes coupling.
\end{abstract}


\input{section/introduction}
\input{section/related_work}
\input{section/pitfall}
\input{section/method}
\input{section/experiment_setup}
\input{section/experiment_result}
\input{section/conclusion.tex}
\section{Acknowledgments}
This work is supported by the National Key Research and Development Project of New Generation Artificial Intelligence, No.:2018AAA0102502; the National Natural Science Foundation of China, No.:U1936203.

\bibliographystyle{IEEETran}
\bibliography{Unified-VCR}

\begin{IEEEbiography}[{\includegraphics[width=1in,height=1.25in,clip,keepaspectratio]{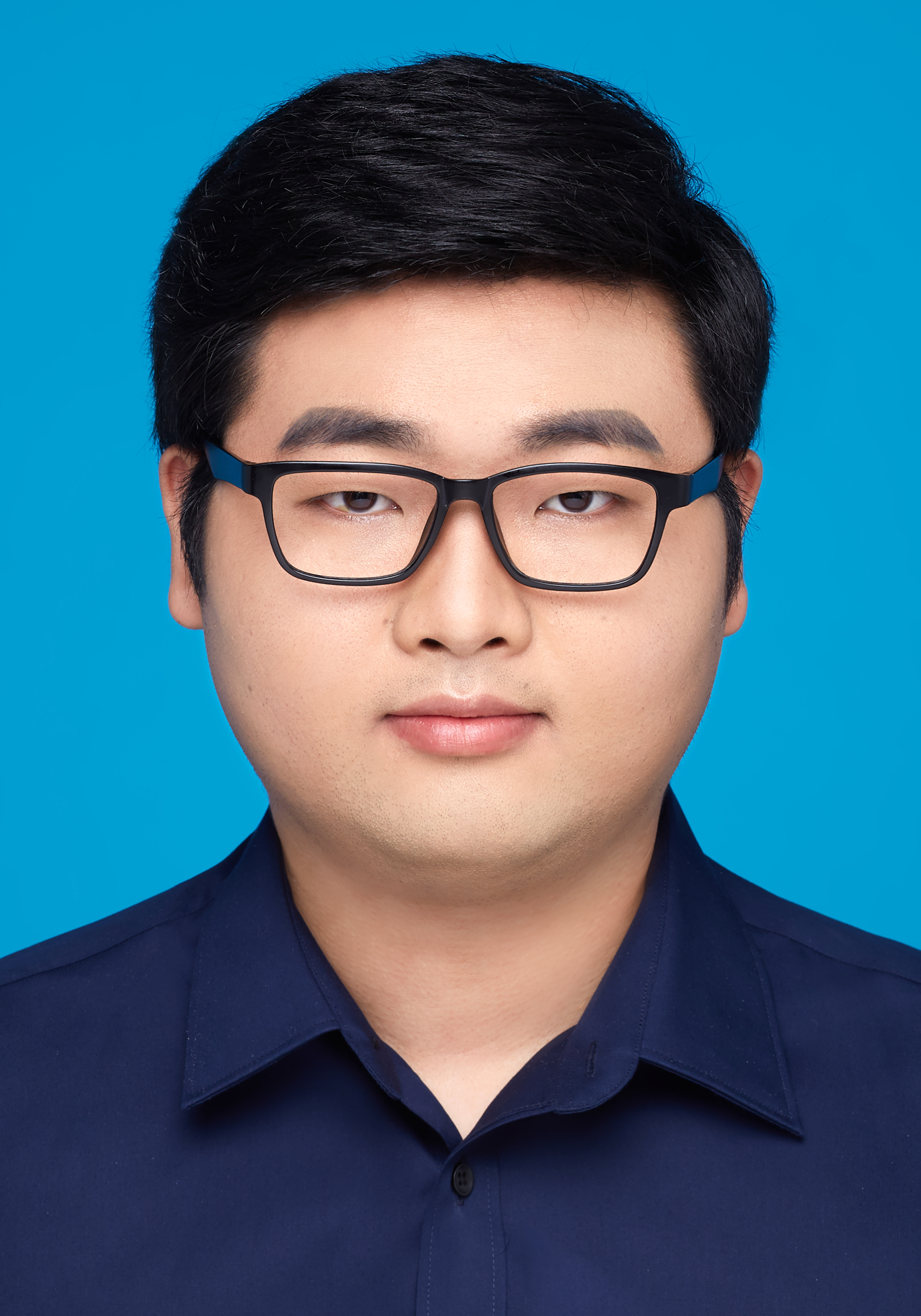}}]{Zhenyang Li}
received the B.Eng. and master's degrees from Shandong University and the University of Chinese Academy of Sciences, respectively. He is currently pursuing the Ph.D. degree at the School of Computer Science and Technology, Shandong University, supervised by Prof. Liqiang Nie. His research interest is multi-modal computing, especially visual question answering.
\end{IEEEbiography}

\begin{IEEEbiography}[{\includegraphics[width=1in,height=1.25in,clip,keepaspectratio]{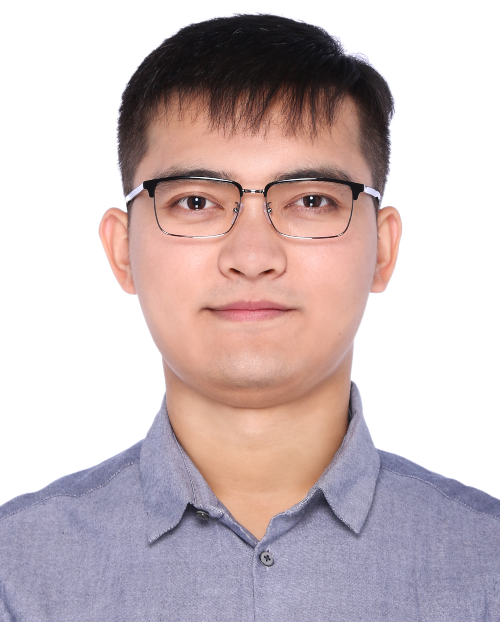}}]{Yangyang Guo}
(Member, IEEE)
is currently a research fellow with the National University of Singapore. 
He has authored or co-authored several papers in top journals, such as IEEE TIP, TMM, TKDE, TNNLS, and ACM TOIS. 
He is a Regular Reviewer for journals, including IEEE TIP, TMM, TKDE, TCSVT; ACM TOIS, and ToMM. 
He was the recipient as an outstanding reviewer for IEEE TMM and WSDM 2022.
\end{IEEEbiography}

\begin{IEEEbiography}[{\includegraphics[width=1in,height=1.25in,clip,keepaspectratio]{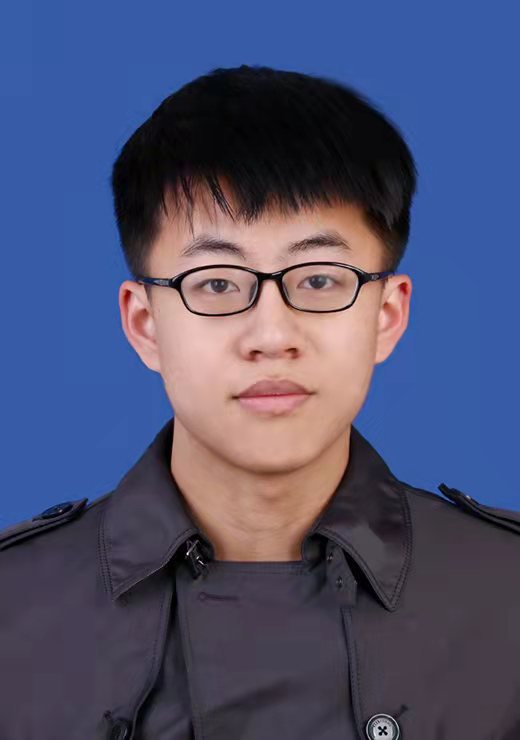}}]{Kejie Wang}
is currently pursuing the B.Eng. degree in computer science from Shandong University. His research interests include visual question answering and computer vision.
\end{IEEEbiography}

\begin{IEEEbiography}[{\includegraphics[width=1in,height=1.25in,clip,keepaspectratio]{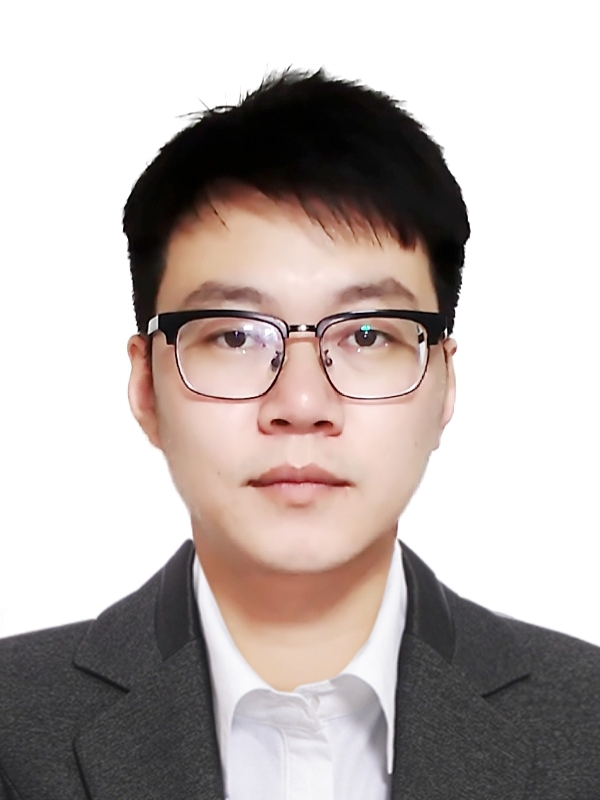}}]{Yinwei Wei}
(Member, IEEE)
received his MS degree from Tianjin University and Ph.D. degree from Shandong University, respectively. Currently, he is a research fellow with NExT, National University of Singapore. His research interests include multimedia computing and recommendation. Several works have been published in top forums, such as ACM MM, IEEE TMM, and TIP. Dr. Wei has served as a PC member for several conferences, such as MM, AAAI, and IJCAI, and the reviewer for TPAMI, TIP, and TMM.
\end{IEEEbiography}

\begin{IEEEbiography}[{\includegraphics[width=1in,height=1.25in,clip,keepaspectratio]{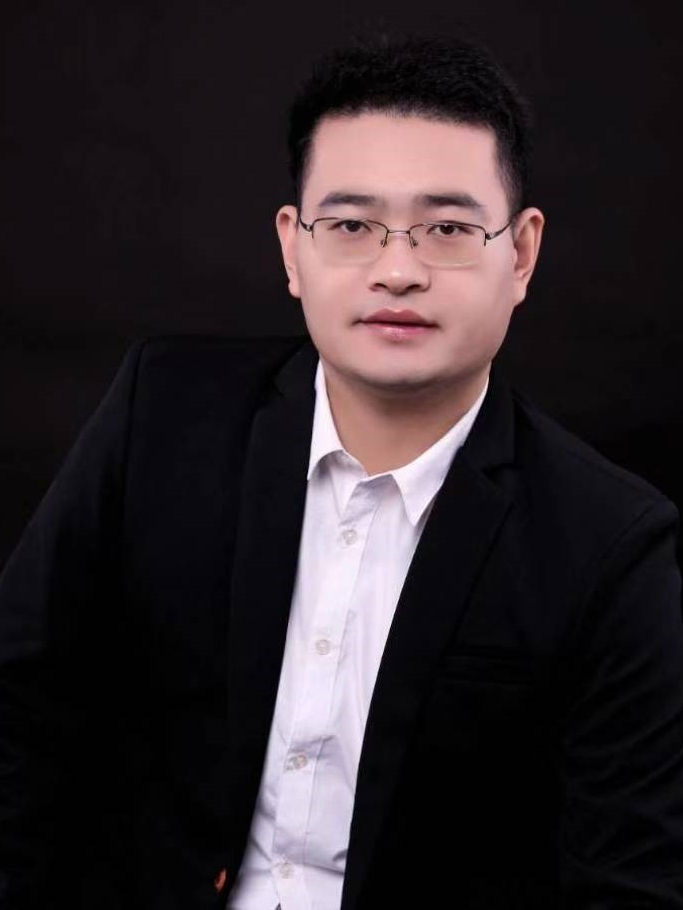}}]{Liqiang Nie}
(Senior Member, IEEE) received the B.Eng. degree from Xi’an Jiaotong University and the Ph.D. degree from the National University of Singapore (NUS). 
He is currently a Professor and the dean of the School of Computer Science and Technology, Harbin Institute of Technology (Shenzhen). 
His research interests lie primarily in multimedia computing and information retrieval.  
He has co-authored more than 200 articles and four books and received more than 14,000 Google Scholar citations. 
He is an AE of IEEE TKDE, IEEE TMM, IEEE TCSVT, ACM ToMM, and Information Science. 
Meanwhile, he is the regular area chair of ACM MM, NeurIPS, IJCAI, and AAAI. 
He is a member of the ICME steering committee. 
He has received many awards, like ACM MM and SIGIR best paper honorable mention in 2019, SIGMM rising star in 2020, TR35 China 2020, DAMO Academy Young Fellow in 2020, and SIGIR best student paper in 2021.
\end{IEEEbiography}

\begin{IEEEbiography}[{\includegraphics[width=1in,height=1.25in,clip,keepaspectratio]{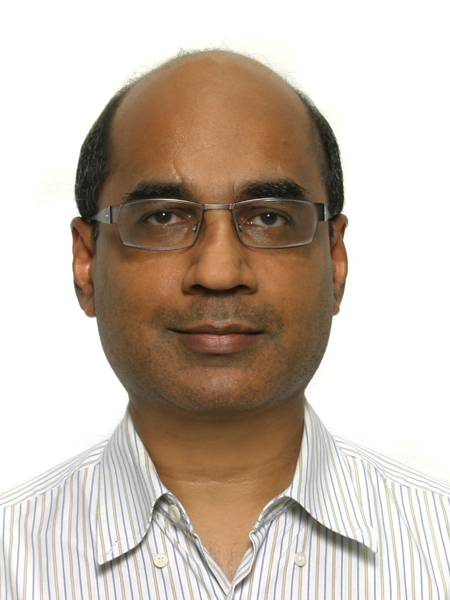}}]{Mohan Kankanhalli}
(Fellow, IEEE)
received the B.Tech. degree from IIT Kharagpur and the M.S. \& Ph.D. degrees from the Rensselaer Polytechnic Institute. 
He is currently the Provost’s Chair Professor at the Department of Computer Science, National University of Singapore. 
He is the Director of N-CRiPT and also the Dean of the School of Computing at the National University of Singapore. 
His current research interests include multimedia computing, multimedia security and privacy, image/video processing, and social media analysis.
\end{IEEEbiography}

\end{document}

%% file: section/introduction.tex
\section{Introduction}
\begin{figure}[t]
  \centering
   \includegraphics[width=0.99\linewidth]{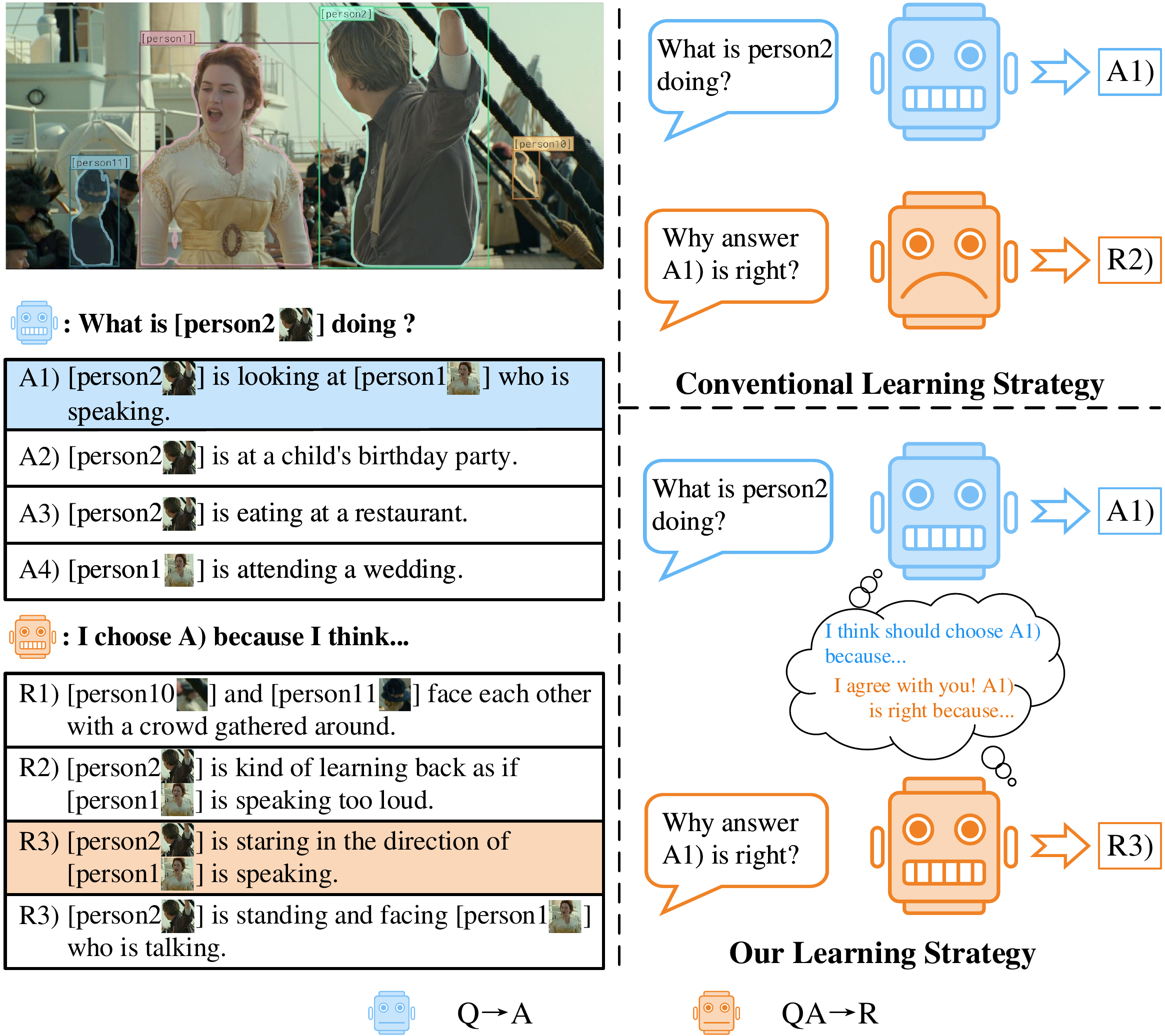}
   \caption{Visual comparison between conventional learning strategy and ours. 
   Unlike previous methods separately treating Q$\rightarrow$A and QA$\rightarrow$R, ours in the bottom, couple these two processes together by their non-separable nature.}
   \label{fig:teaser}
\end{figure}

\IEEEPARstart{V}{isual} Question Answering (VQA) is to answer a natural language question pertaining to a given image~\cite{VQA1, VQA2}. 
Despite its noticeable progress, existing VQA benchmarks merely address simple recognition questions (\eg \emph{how many} or \emph{what color}), while neglecting the explanation of answering prediction. 
In light of this, the task of Visual Commonsense Reasoning (VCR)~\cite{R2C} has recently been introduced to bridge this gap.
Beyond answering the cognition-level questions (Q$\rightarrow$A) as canonical VQA does, VCR further prompts to predict a rationale for the right answer option (QA$\rightarrow$R), as shown in Figure \ref{fig:teaser}.
z

VCR is made more challenging than VQA roughly in terms of the following two aspects: 
1) On the data side -- 
The images in the VCR dataset describe more sophisticated scenes (\eg social interactions or mental states). 
Therefore, the collected questions are rather difficult and often demand high-level visual reasoning capabilities (\eg \emph{why} or \emph{how}).
And 2) on the task side -- It is difficult holistically predict both the right answer and rationale. 
VCR models should first predict the right answer, based on which the acceptable rationale can be further inferred from a few candidates.
As question answering about the complex scenes has been proved non-trivial, inferring the right rationale simultaneously in VCR thus leads to more difficulties.

In order to address the above research challenges, several prior efforts have been dedicated to VCR over the past few years. 
The initial attempts contribute to designing specific architectures to approach VCR, such as R2C~\cite{R2C} and CCN~\cite{CCN}. 
In addition, recent BERT-based pre-training approaches, such as ViLBERT~\cite{ViLBERT} and ERNIE-ViL~\cite{ERNIE-ViL}, are reckoned as a better solution to the first challenge. 
Specifically, a \emph{pretrain-then-finetune} learning scheme is adopted to transfer knowledge from large-scale vision-language datasets~\cite{Conceptual_Captions} to VCR.
These methods, though keep advancing numbers of the benchmark, all consider Q$\rightarrow$A and QA$\rightarrow$R as two independent processes. 
As a result, the question answering and rationale inference processes are far from being resolved. 
Namely, the second challenge still remains unsettled in literature.

Separately treating Q$\rightarrow$A and QA$\rightarrow$R brings adverse effects on visual reasoning, considering that these two processes share a consistent goal by nature. 
On the one hand, Q$\rightarrow$A entails the reasoning clues for QA$\rightarrow$R to infer the right rationale. 
On the other hand, QA$\rightarrow$R offers essential explanations for Q$\rightarrow$A to justify why the predicted answer is correct.
Nevertheless, separately treating these two processes makes VCR degenerate into two independent VQA tasks\footnote{Note that for QA$\rightarrow$R, the \emph{query} to a VQA model now becomes the concatenation of the original question and the right answer. 
And the corresponding \emph{answer} choices are the set of candidate rationales.} 
As a result, QA$\rightarrow$R has to rely on other unexpected information (\eg word overlaps between answers and rationales), rendering the explanation less meaningful, due to the absence of guidance from Q$\rightarrow$A.
To bring this problem explicit to readers, we perform extensive in-depth empirical experiments in this work (ref. Section \ref{sec: section3}).

In addition, by conforming to the task intuition and human cognition, the answering (Q$\rightarrow$A) and reasoning (QA$\rightarrow$R) should be made cohesive and consistent rather than separate. 
To achieve this goal, we propose a novel plug-and-play knowledge distillation enhanced framework to perform Answering and Reasoning Coupling (\textbf{ARC} for short). 
The key to our framework is the introduction of another proxy process \ie QR$\rightarrow$A. 
It targets answer prediction with the inputs of the given question and the right rationale. 
The intuition is that QR$\rightarrow$A shares the same goal with Q$\rightarrow$A yet is more information-abundant, and its reasoning procedure should be similar with QA$\rightarrow$R as these two are actually complementary. 
In our implementation, we devise two novel Knowledge Distillation (KD) modules: KD-A and KD-R. The former aligns the predicted logits between Q$\rightarrow$A and QR$\rightarrow$A since answering with the right rationale is expected to be more confident. 
While the latter, which is pivotal to maintaining semantic consistency between QA$\rightarrow$R and QR$\rightarrow$A, aligns the fused multi-modal features between them. 
With the aid of the above two KD modules, the models are enabled to couple the answering and reasoning together, making visual reasoning more faithful. 

In summary, the contribution of this paper is threefold:
\begin{itemize}
\item We revisit VCR from the perspective of coupling the Q$\rightarrow$A and QA$\rightarrow$R processes. 
With extensive probing tests, we find that separately treating the two processes, as adopted by existing VCR methods, is detrimental to visual reasoning. 
To the best of our knowledge, this work is among the first efforts to jointly explore the two processes in VCR.
\item We propose a novel plug-and-play knowledge distillation enhanced framework. 
Our newly introduced QR$\rightarrow$A branch serves as a proxy to efficiently couple Q$\rightarrow$A and QA$\rightarrow$R.
\item We apply this framework to five baseline methods and conduct experiments on the VCR benchmark dataset. 
The results demonstrate both the effectiveness and generalization capability of our proposed framework. 
As a side product, the code has been released to facilitate other researchers\footnote{https://github.com/SDLZY/ARC.}.
\end{itemize}

The rest of this paper is structured as follows. 
We briefly review the related literature in Section \ref{sec: related work} and then discuss the pitfalls of existing methods in Section \ref{sec: section3}, 
followed by our proposed ARC framework in Section \ref{sec: method}. 
Experimental setup and results analyses are presented in Section \ref{sec: experimental setup} and Section \ref{sec: experimental results}, respectively. 
We finally conclude our work and outline the future work in Section \ref{sec: conclusion}.

%% file: section/related_work.tex
\section{Related Work}
\label{sec: related work}
\subsection{Explanation in Visual Question Answering}
Traditional VQA models often adopt a CNN-RNN learning paradigm, wherein the images are encoded via a pre-trained CNN network, in parallel to an RNN network taking the questions as input~\cite{VQA3, VQA1, VQA4}. 
Broad efforts have been devoted to applying the attention~\cite{BUTD, Re-attention}, modular structure~\cite{NMN, Clevr}, or external knowledge~\cite{Fvqa, Ok-vqa} to VQA models.
Recently, some researchers have recognized the unfaithfulness of existing approaches. 
For example, \cite{HINT, Adavqa, Dont_just_assume} aim to identify and overcome the language prior problem, \ie the answers are blindly predicted based on textual shortcuts between questions and answers. 
Other methods endeavor to introduce explanation for VQA models~\cite{VQA-E, Explanation_vs_attention}.
For instance,  visual explanation approaches often harness the heat map to achieve \emph{where to look}, such as Gram-CAM~\cite{Grad-cam} or U-CAM~\cite{U-cam}. 
VQA-HAT~\cite{VQA-HAT} collects human attention data which are utilized to supervise visual attention learning. 
HINT~\cite{HINT} encourages VQA models to be sensitive to the same input regions as humans. 
Different from these approaches, VQA-E~\cite{VQA-E} requires a VQA model to generate a textual explanation for the answer prediction.
As a natural extension, both the visual and textual explanations are also well studied~\cite{VQA-X, VQA5}.

\subsection{Visual Commonsense Reasoning}
VCR contributes to an indispensable branch of the explainable VQA~\cite{R2C}. 
It involves two processes: question answering (Q$\rightarrow$A) and rationale inference (QA$\rightarrow$R), whereby both are embodied in a multiple-choice fashion. 
VCR instructs an approach to not only answer the given question but also provide a reasonable explanation to Q$\rightarrow$A. 
Accordingly, the answer prediction is made more transparent and user-friendly to humans.

Existing studies are in fact sparse due to its challenging nature. 
Some of them resort to designing specific model architectures~\cite{R2C, HGL, CCN, ECMR}. 
For instance, R2C~\cite{R2C} adopts a three-step fashion, \emph{grounding} texts with respect to the involved objects; \emph{contextualizing} the answer with the corresponding question and objects; and \emph{reasoning} over the shared representation.
Inspired by the neuron connectivity of brains, CCN~\cite{CCN} dynamically models the visual neuron connectivity, which is contextualized by the queries and responses.
HGL~\cite{HGL} leverages a vision-to-answer and a dual question-to-answer heterogeneous graphs to seamlessly bridge vision and language.
Following approaches explored more fine-grained cues, such as ECMR~\cite{ECMR} incorporating the syntactic information into visual reasoning~\cite{R3_5_ref2}, MLCC~\cite{R3_MM} using counterfactual thinking to generate informative samples.

Thereafter, BERT-based pre-training approaches have been extensively explored in the vision-and-language domain. 
Most of these methods employ a \emph{pretrain-then-finetune} scheme and achieve significant performance improvement on VQA benchmarks including VCR~\cite{ViLBERT, UNITER, VL-BERT}.
The models are often first pre-trained on large-scale vision-and-language datasets (such as Conceptual Captions~\cite{Conceptual_Captions}), and then fine-tuned on the downstream VCR. 
For instance, some models are pre-trained on image-text datasets to align the visual-linguistic clues through single-stream architectures (\eg VL-BERT~\cite{VL-BERT}, UNITER~\cite{UNITER}, 12-in-1~\cite{12-in-1}).
MERLOT RESERVE~\cite{MERLOT_RESERVE} introduces more modalities, \ie sound and video, into cross-modal pre-training, and significantly outperforms other methods.

However, one limitation still prevents these methods from further advancement, namely, the Q$\rightarrow$A and QA$\rightarrow$R processes are tackled independently. 
Such a strategy makes the answer prediction and rationale inference become two independent VQA tasks.
In this work, we propose to address this issue by combining these two processes. 

\subsection{Knowledge Distillation}\label{sec:KD}
The past few years have witnessed the noticeable development of KD~\cite{KD, Lipschitz}. 
As an effective model compression tool, KD transfers knowledge from a cumbersome large network (teacher) to a compacter network (student). 
Based on its application scope, previous KD methods can be roughly categorized into two groups -- logit-based and feature-based. 
The logit-based methods~\cite{KD, KD_Logits} encourage the student to imitate the output from a teacher model. 
For example, the vanilla KD utilizes the softened logits from a pre-trained teacher network as extra supervision to instruct the student~\cite{KD}. 
In contrast, feature-based methods~\cite{KTAN, MEAL, Lipschitz} attempt to transfer knowledge in intermediate features between the two networks. 
FitNet~\cite{FitNets} directly aligns the embeddings of each input. 
Attention Transfer~\cite{Paying_More_Attention_to_Attention} extends FitNet from embeddings to attention matrices for different levels of feature maps. 
To close the performance gap between the teacher and student, RCO~\cite{RCO} presents route-constrained hint learning, which is employed to supervise the student by the outputs of hint layers from the teacher.
Besides, FSP~\cite{FSP} estimates the Gram matrix across layers to reflect the data flow of how the teacher network learns.

%% file: section/pitfall.tex
\section{Pitfall of Existing VCR Methods}
\label{sec: section3}
VCR is challenging due to the fact that the model is required to not only answer a question but also reason about the answer prediction. 
Accordingly, the answering and reasoning processes are complementary and inseparable from each other. 
However, existing methods often treat Q$\rightarrow$A and QA$\rightarrow$R separately, resulting in sub-optimal visual reasoning as revealed in the following two aspects.

\subsection{Language Shortcut Recognition}
\label{sec: shorcuts}
Besides the base Q$\rightarrow$A model, representative approaches mostly adopt another independent model for QA$\rightarrow$R.
Since there is no connection between these two processes, we hereby raise a question:
what kinds of clues do these methods employ, excluding the Q$\rightarrow$A reasoning information? 
With this concern, at the first step, we recognize that \emph{the overlapped words between right answers and rationales dominate the rationale inference}. 
For instance, in Figure~\ref{fig:teaser}, the correct rationale largely overlaps with the right answer, \eg the `[person1]' and `[person2]' tag, and the word `speaking'. 
This may lead the models to predict rationale based on these shortcuts rather than performing visual explanation for Q$\rightarrow$A~\cite{Adavqa, rescale-prior}. 
The evidence is elaborated as follows.

\textbf{QA$\rightarrow$R Performance \emph{w/o} Q}. 
We input only the correct answer as the query to predict rationales with other settings untouched and show the results in Table \ref{tab:a2r}. 
One can observe that the three models only degrade slightly when removing the question input. 
As VCR is a question-driven task, visual reasoning becomes meaningless under such input removal conditions. 

\begin{table}[h]
  \centering
  \caption{Rationale prediction performance comparison with and without questions}.\label{tab:a2r}
  \scalebox{1}{
  \begin{tabular}{lcc}
    \toprule
    Model           & QA$\rightarrow${R}    & A$\rightarrow${R}\\
    \midrule
    R2C ~\cite{R2C} & 67.2                  & 66.3\\
    HGL ~\cite{HGL} & 70.6                  & 69.8\\
    CCN ~\cite{CCN} & 70.5                  & 70.4\\
    \bottomrule
  \end{tabular}}
\end{table}

\begin{figure}[h]
  \centering
   \includegraphics[width=0.85\linewidth]{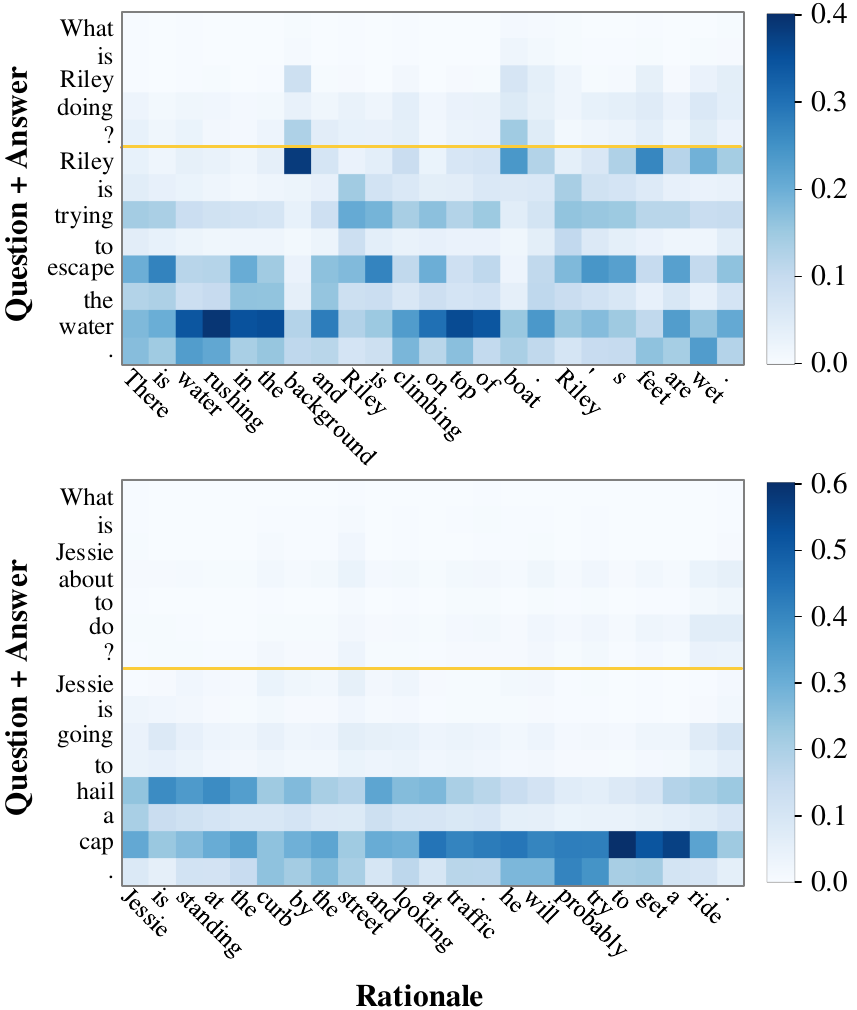}

   \caption{Illustration of the attention distribution of QA$\rightarrow$R from R2C. 
   The orange line splits the question and the answer from the query.}
   \label{fig:figure3}
\end{figure}

\textbf{Attention Distribution over Queries}. 
As attention plays an essential part in current VCR models, we then design experiments to explore the shortcut from this perspective.
In this experiment, we consider only the QA$\rightarrow$R. 
Given the attention map $\mathbf{W}\in{\mathbb{R}^{(l_{q}+l_{a})\times{l_r}}}$ for a query-rationale, wherein the query is the original question appended with the right answer. 
pair, where $l_q$, $l_a$, and $l_r$ respectively denote the length of the question, answer, and rationale, we calculate the attention contribution from the answer side only and analyze below.
Specifically, we find that the median attention value obtained from answers in three methods (HGL, R2C, and CCN) on the validation set is 0.72, 0.78, and 0.86, respectively, indicating that the these models pay more attention to answers rather than the holistic question-answer inputs. 
In this way, they mainly rely on the answer information to predict the right rationale, while the questions are somewhat ignored.
Two examples produced by R2C are illustrated in Figure \ref{fig:figure3}. 

\subsection{Generalization on Out-of-Domain Data}
\label{sec:overfit}
In addition to the shortcut problem, thereafter, we also study the generalization capability of existing methods on the out-of-domain data. 
To implement this, we first rewrite some sentences (including questions, answers, and rationales) while maintaining their semantics. 
The Paraphrase-Online tool\footnote{https://www.paraphrase-online.com/.} is leveraged to achieve this, whereby we substitute some verbs/nouns with synonyms. 
We show some examples in Table \ref{tab:paraphrase_example}.
\begin{table}[!t]
  \centering
  \caption{Rewritten examples from our experiments.}
  \label{tab:paraphrase_example}
  \scalebox{1}{
  \begin{tabular}{p{4cm}|p{4cm}}
    \toprule
    Original                                                                  & Paraphrased \\
    \midrule
    What is \emph{[person4]} getting ready to do?                             & What is \emph{[person4]} getting prepared to do? \\
    \midrule
    He's about to start singing.                                              & He's around to begin singing. \\
    \midrule
    Everyone else is singing and he seems to be getting ready to follow suit. & Everyone else is singing and he appears to be getting prepared to copy them.\\
    \bottomrule
  \end{tabular}}
\end{table}
In the next step, we apply the above methods to the rewritten data and evaluate their performance. 
We also test the model performance with further BERT pre-training on the rewritten textual data.
From the results in Table \ref{tab:origin_vs_paraphrase}, we find that all models degrade drastically on these new instances, implying that they largely overfit on the in-domain data while lack generalization ability on the out-of-domain ones.

\begin{table}[h]
  \centering
  \caption{Model performance comparison over two versions of the validation set. 
  \emph{Origin} and \emph{Rewritten} denote the models are tested on the original and rewritten VCR validation set, respectively.
  LP represents further language pre-training on the rewritten textual data.
  \label{tab:origin_vs_paraphrase}}
  \scalebox{0.92}{
  \begin{tabular}{r|ccc|ccc}
    \toprule
    \multirow{3}*{Model}    & \multicolumn{3}{c|}{Q$\rightarrow$A}                                          & \multicolumn{3}{c}{QA$\rightarrow$R}\\
                            \cmidrule(lr){2-4}                                                              \cmidrule(lr){5-7}
                            & \multirow{2}*{Origin}         & \multicolumn{2}{c|}{Rewritten}                & \multirow{2}*{Origin}         & \multicolumn{2}{c}{Rewritten}\\
                                                            \cmidrule(lr){3-4}                                                              \cmidrule(lr){6-7}
                            &                               & \emph{w/} LP     & \emph{w/o} LP              &                               & \emph{w/} LP     & \emph{w/o} LP\\
    \midrule
    R2C~\cite{R2C}          & 63.8                          & 55.2             & 52.8                       & 67.2                          & 59.9       & 54.7\\
    CCN~\cite{CCN}          & 66.8                          & 55.3             & 52.6                       & 70.6                          & 60.3       & 56.5\\
    TAB-VCR~\cite{TAB-VCR}  & 69.9                          & 60.3             & 56.4                       & 72.2                          & 66.1       & 60.5\\
    \bottomrule
  \end{tabular}}
\end{table}

Based on these two findings, we notice that the current models fail to conduct visual reasoning based on the answering clues. 
They instead, leverage superficial correlations to infer rationale or simply overfit on the in-domain data. 
To approach these problems, a VCR model is expected to benefit from the joint training of answering and reasoning.

%% file: section/method.tex
\section{Method}
\label{sec: method}
The above probing tests demonstrate that separately treating the two processes in VCR leads to unsatisfactory outcomes. 
To overcome this, we propose an ARC framework to couple Q$\rightarrow$A and QA$\rightarrow$R together.
In this work, we introduce another new branch, namely QR$\rightarrow$A, as a bridge to achieve this goal.

\subsection{Preliminary}
Before delving into the details of our ARC, we first outline the three processes involved in our framework and their corresponding learning functions.
\subsubsection{Process Definition}
\label{sec: process deifne}
Our framework contains three processes: two of them are formally defined by the original VCR~\cite{R2C} (\ie Q$\rightarrow$A and QA$\rightarrow$R) and the other is a new process introduced by this work (QR$\rightarrow$A). 
Note that all three processes are formulated in a multiple-choice format. 
That is, given an image $I$ and a \emph{query} related to the image, the model is required to select the right option from \emph{several candidate responses}. 

\textbf{Q$\rightarrow$A}: The query is embodied with a question $Q$, and the candidate responses are a set of answer choices $\mathcal{A}$. 
The objective is to select the right answer $A_{+}$,

\begin{equation}
\label{eq: objective_q2a}
    A_{+} = \mathop{\arg\max}\limits_{A_i\in{\mathcal{A}}} {f^{A}(A_i \mid Q, I)},
\end{equation}
where $f^{A}$ denotes the Q$\rightarrow$A model.

\textbf{QA$\rightarrow$R}: The query is the concatenation of the question $Q$ and the right answer $A_{+}$. 
A set of rationales $\mathcal{R}$ constitutes the candidate responses and the model is expected to choose the right rationale $R_{+}$,
\begin{equation}
\label{eq: objective_qa2r}
    R_{+} = \mathop{\arg\max}\limits_{R_i\in{\mathcal{R}}} {f^{R}(R_i \mid Q, I, A_+)},
\end{equation}
where $f^{R}$ denotes the QA$\rightarrow$R model.

One can see that it is difficult to directly connect these two since the involved parameters are not shared and the input to QA$\rightarrow$R includes the ground-truth answer rather than the predicted one. 
In view of this, to bridge these two, we introduce the QR$\rightarrow$A as a proxy.

\textbf{QR$\rightarrow$A}: The query and responses are respectively the concatenation of the question $Q$ along with the right rationale $R_{+}$ and answer choices. The objective is,
\begin{equation}
    A_{+} = \mathop{\arg\max}\limits_{A_i\in{\mathcal{A}}} {f^{C}(A_i \mid Q, I, R_{+})},
\end{equation}
where $f^{C}$ denotes the QR$\rightarrow$A model.
On the one hand, QR$\rightarrow$A shares the consistent objective with Q$\rightarrow$A. 
On the other hand, its reasoning should be similar to that of QA$\rightarrow$R. 
These two factors make QR$\rightarrow$A a good proxy for connecting  Q$\rightarrow$A and QA$\rightarrow$R.

\subsubsection{Training Pipeline}
\label{sec: classical learning strategy}
In general, for the feature extractors, the VCR model often uses a pre-trained CNN network to obtain the visual features from the input image $I$, and an RNN-based or Transformer-based model to extract the textual features of the query and response.
Thereafter, a multi-modal fusion module is employed to obtain the joint representation, followed by the classifier to predict the logit $\Tilde{y}_i$ for the response $i$.

To achieve the objectives in Equation~\ref{eq: objective_q2a} and \ref{eq: objective_qa2r}, previous methods often separately optimize the following two cross-entropy losses,
\begin{equation}
\begin{cases}
    \mathcal{L}^{A} = -\sum\nolimits^{|\mathcal{A}|}_i y_i^A\log\frac{\exp{{\Tilde{y}_i^A}}}{\sum\nolimits_j\exp{{\Tilde{y}_j^A}}}, \\
    \mathcal{L}^{R} = -\sum\nolimits^{|\mathcal{R}|}_i y_i^R\log\frac{\exp{{\Tilde{y}_i^R}}}{\sum\nolimits_j\exp{{\Tilde{y}_j^R}}},
\end{cases}
\end{equation}
where $y_i^A$ and $y_i^R$ denote the ground truth label of answer $A_i$ and rationale $R_i$, respectively.

In existing VCR methods, models $f^A$ and $f^R$ often share identical architecture and are trained separately. 
As a result, the two processes are reduced into two independent VQA tasks, resulting in connection absence between answering and reasoning. 
In the next, we present our model-agnostic framework to couple the two successive processes, \ie Q$\rightarrow$A and QA$\rightarrow$R, together. 
\begin{figure*}
  \centering
  \includegraphics[width=1.0\linewidth]{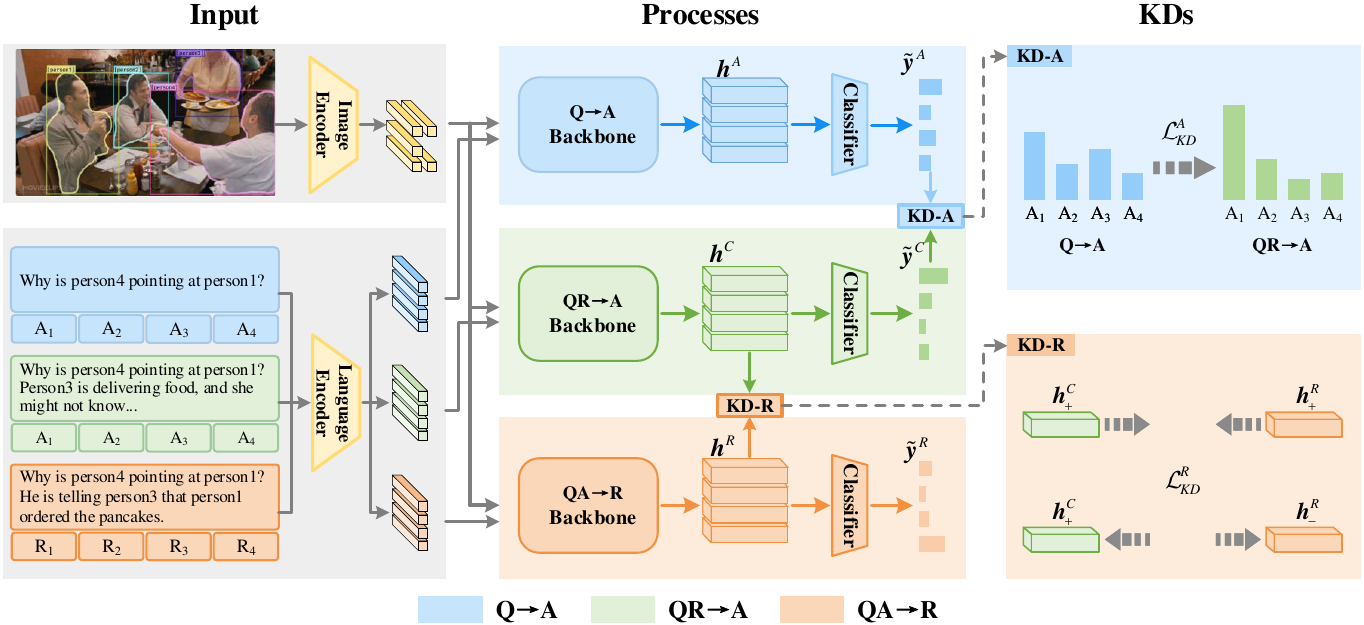}
  \caption{The overall architecture of our proposed framework. 
  There are three processes, \ie Q$\rightarrow$A, QR$\rightarrow$A, and QA$\rightarrow$R. 
  We bridge these processes with two KD modules: 
  KD-A is leveraged to align the predicted logits between QR$\rightarrow$A and Q$\rightarrow$A; 
  and KD-R aims to maintain semantic consistency between QR$\rightarrow$A and QA$\rightarrow$R via the feature-level knowledge distillation operation.}\label{fig:framework}
\end{figure*}

\subsection{Proposed Method}
In order to effectively integrate the above three processes, we propose a dual-level knowledge distillation framework and illustrate it in Figure~\ref{fig:framework}. 
Specifically, after extracting the fused multi-modal features with the aforementioned backbones, two parallel KD modules are introduced to bridge the QR$\rightarrow$A with the other two processes. 

Our first KD module is leveraged to align the predicted scores from QR$\rightarrow$A and Q$\rightarrow$A. 
The two processes share the same objective and QR$\rightarrow$A offers ample information for picking the right answer. 
In this way, we take the predictions from QR$\rightarrow$A as the teacher and the predictions from Q$\rightarrow$A as the student. 
Our second KD module is required to align the feature learning between QR$\rightarrow$A and QA$\rightarrow$R. Despite the objectives being different, the reasoning of these two is actually similar.
That is, answering with the given right rationale is simply the reverse of reasoning given the right answer, which allows these two processes to share akin features.

\subsubsection{Knowledge Distillation between QR$\rightarrow$A and Q$\rightarrow$A (KD-A)}
\label{sec: kd1}
Since the rationale is predicted for explaining the right answer, we thus empirically leverage the other direction, \ie incorporating the correct rationale with the given question for answer prediction. 
It is intuitive that when combined with the correct rationale, the answering confidence should be enhanced.
In light of this, we simply take the output logits from QR$\rightarrow$A as the teacher and use them guide the learning of Q$\rightarrow$A.

Specifically, in this KD module, the knowledge is encoded and transferred in the form of softened scores: 
it aligns the output probability $\boldsymbol{p}^A$ (from the student $f^A$) with the $\boldsymbol{p}^C$ (from the teacher $f^C$). 
To avoid the over-confidence problem~\cite{KD_Logits}, a relaxation temperature $T > 1$ is introduced to soften the logits $\boldsymbol{\Tilde{y}}^C$ from $f^C$. 
The same relaxation is applied to the output logits $\boldsymbol{\Tilde{y}}^A$ of $f^A$,
\begin{equation}
\begin{cases}
    p^C_i = \frac{\exp{(\Tilde{y}^C_i/T)}}{\sum\nolimits_{j=1}^{|\mathcal{A}|}\exp{(\Tilde{y}^C_j/T)}}, \\
    p^A_i = \frac{\exp{(\Tilde{y}^A_i/T)}}{\sum\nolimits_{j=1}^{|\mathcal{A}|}\exp{(\Tilde{y}^A_j/T)}}.
\end{cases}
\end{equation}
We then employ the KL divergence to align the predicted scores as follows,
\begin{equation}
\label{eq: kd1}
\begin{split}
    \mathcal{L}_{KD}^A &= D_{KL}(\boldsymbol{p}^C||\boldsymbol{p}^A).
\end{split}
\end{equation}

\subsubsection{Knowledge Distillation between QR$\rightarrow$A and QA$\rightarrow$R (KD-R)}
We argue that the features learned by QR$\rightarrow$A are more reliable than that of QA$\rightarrow$R. 
The support for this comes from two aspects: QR$\rightarrow$A resembles \emph{raising questions with right answers}, providing explicit evidence for feature learning; 
leveraging QA$\rightarrow$R to perform reasoning might lead to the textual shortcut problem, as discussed in Section~\ref{sec: shorcuts}.

To this end, we design another KD module to align the feature learning between QR$\rightarrow$A and QA$\rightarrow$R, which aims to maintain semantic consistency.
The intermediate features that are directly inputted to the final classifier are taken as a proxy.
Specifically, we obtain one teacher feature $\boldsymbol{h}^C_{+}$ from the teacher model $f^{C}$; 
and several student features $\boldsymbol{h}^R_i$ from the student model $f^{R}$ wherein only one is deemed as positive -- $\boldsymbol{h}^R_+$ according to the ground-truth rationale $R_+$ while others are all negative $\boldsymbol{h}^R_-$.
Thereafter, we estimate the similarity score with the following formula,
\begin{equation}
    s_i=( \boldsymbol{h}^C_+)^T \boldsymbol{h}^R_i,
\end{equation}
Especially, $s_+$ should be larger than $s_-$ in optimal condition.
To enable the training of this module, we adopt the rationale labels as supervision and perform KD-R with,
\begin{equation}
\label{eq: kd2}
    \mathcal{L}^{R}_{KD} = -\sum^{|\mathcal{R}|}_i y_i^R\log\frac{\exp{s_i}}{\sum\nolimits_j^{|\mathcal{R}|}\exp{s_i}}.
\end{equation}

\subsection{Training Protocol}
With the above two KD modules, we then combine them with the original answering and reasoning losses as follows,
\begin{equation}
\label{eq: q2a}
    \mathcal{L}_{Q\rightarrow{A}} = \alpha{\mathcal{L}_{KD}^A} + (1-\alpha)\mathcal{L}^A,
\end{equation}
\begin{equation}
\label{eq: qa2r}
    \mathcal{L}_{QA\rightarrow{R}} = \beta{\mathcal{L}_{KD}^R} + (1-\beta)\mathcal{L}^R,
\end{equation}
where $\alpha$ and $\beta$ serve as hyper-parameters to balance the two KD losses.

Finally, we train the entire framework in an end-to-end fashion,
\begin{equation}
\label{loss: all}
    \mathcal{L} = \mathcal{L}_{Q\rightarrow{A}} + \mathcal{L}_{QA\rightarrow{R}}.
\end{equation}
Different from previous methods that output answers and rationales separately, in our framework, the answers and rationales are predicted at the same time. 
In this way, the Q$\rightarrow$A and QA$\rightarrow$R are made more cohesive than these methods.

The optimization procedure of our proposed ARC, as shown in Algorithm~\ref{code:recentEnd}, contains two stages. 
In the first stage, we train the teacher network $f^C$ until its cross-entropy loss converges (line 1-4),
\begin{equation}
\label{eq: ce_C}
    \mathcal{L}^{C} = -\sum^{|\mathcal{A}|}_i y_i^A\log\frac{\exp{{\Tilde{y}_i^C}}}{\sum\nolimits_j\exp{{\Tilde{y}_j^C}}},
\end{equation}
where $\Tilde{y}_i^C$ denotes the output of $f^C$ corresponding to answer $A_i$. 
Secondly, $f^A$ and $f^R$ are optimized together, where the cross-entropy losses and KD losses are both considered (line 5-10). 
During inference, the $f^C$ branch is removed, and only $f^A$ and $f^R$ are responsible to predict answers and rationales, respectively.
\begin{algorithm}[h]  
  \renewcommand{\algorithmicrequire}{\textbf{Input:}}
  \renewcommand{\algorithmicensure}{\textbf{Output:}}
  \caption{Training Procedure of ARC} \label{code:recentEnd}
  \begin{algorithmic}[1]
    \Require
    {Training set $\mathcal{D}^A$, $\mathcal{D}^R$ and $\mathcal{D}^C$ for three processes respectively.}
    
    \Repeat
      \State Randomly select a mini-batch $\mathcal{B}^C\in\mathcal{D}^C$;
      \State Employ the teacher model $f^C$ on $\mathcal{B}^C$ and compute $\mathcal{L}^C$ by Equation~\ref{eq: ce_C};
      \State Update weights in $f^C$\;
    \Until{$f^C$ converges}

    \Repeat
      \State Randomly select the mini-batches $\mathcal{B}^A\in\mathcal{D}^A$, $\mathcal{B}^R\in\mathcal{D}^R$ and $\mathcal{B}^C\in\mathcal{D}^C$;
      \State Employ $y^A$, $y^R$ and $y^C$ on $\mathcal{B}^A$, $\mathcal{B}^R$ and $\mathcal{B}^C$, respectively;
      \State Compute $\mathcal{L}_{Q\rightarrow{A}}$ by Equation~\ref{eq: q2a} to distil knowledge from $y^C$ to $y^A$;
      \State Compute $\mathcal{L}_{QA\rightarrow{R}}$ by Equation~\ref{eq: qa2r} to distil knowledge from $y^C$ to $y^R$;
      \State Sum above losses by Equation~\ref{loss: all};
      \State Update weights in $f^A$ and $f^R$\;
    \Until{both $f^A$ and $f^R$ converge}
    
    \Ensure The trained student model $f^A$ and $f^R$.
  \end{algorithmic}
\end{algorithm} 

%% file: section/experiment_setup.tex
\section{Experimental Setup}
\label{sec: experimental setup}
\subsection{Datasets and Evaluation Protocols}
We conducted extensive experiments on the VCR~\cite{R2C} benchmark dataset.
Specifically, the images are extracted from the movie clips in LSMDC~\cite{LSMDC} and MovieClips\footnote{youtube.com/user/movieclips.}, wherein the objects in images are detected by an Mask-RCNN model~\cite{MASK_RCNN}.
We utilized the official dataset split, and the number of instances for training, validation, and test sets are 212,923, 26,534, and 25,263, respectively. 
For Q$\rightarrow$A and QA$\rightarrow$R, four options serve as candidates and only one is correct. 

Regarding the evaluation metric, we adopted the popular accuracy metric for Q$\rightarrow$A, QA$\rightarrow$R, and Q$\rightarrow$AR. 
For Q$\rightarrow$AR, the prediction is \textbf{right only when both the answer and rationale are selected correctly}. 
Since the test set labels are not available, we reported the performance of our best model on the test set once and performed other experiments on the validation set.

The PyTorch toolkit is leveraged to implement our method and all the experiments were conducted on a single GeForce RTX 2080 Ti GPU.
We strictly followed the training and inference protocols of every baseline. 
We employed grid-search to decide the optimal value of the hyper-parameters of $\alpha$ and $\beta$ in Equation~\ref{eq: q2a} and~\ref{eq: qa2r}, respectively.
Both parameters are tuned in the range of [0.0, 1.0] with a step size of 0.2.

\subsection{Baselines}
We compared our method with the following two sets of baselines. 
\subsubsection{VQA baselines} RevisitedVQA~\cite{RevisitedVQA}, BottomUpTopDown~\cite{BUTD}, MLB~\cite{MLB} and MUTAN~\cite{MUTAN}. 
These methods are originally developed for VQA and adapted for VCR in this paper.

\subsubsection{Methods specifically designed for VCR}
We employed our method to the following baselines to validate its effectiveness: 
Traditional VCR models -- R2C~\cite{R2C}, CCN~\cite{CCN}, andTAB-VCR~\cite{TAB-VCR};
and recent VL-Transformers -- UNITER~\cite{UNITER} and VL-BERT~\cite{VL-BERT}.

%% file: section/experiment_result.tex
\section{Experimental Results}
\label{sec: experimental results}
\subsection{Overall Performance Comparison}
\begin{table}[!t]
\centering
  \caption{Performance comparison on VCR validation and testing sets.}
  \label{tab:overall}
  \scalebox{1}{
  \begin{tabular}{p{1pt} p{2.1cm} | c c | c c | c c}
    \toprule
    \multicolumn{1}{c}{}&\multicolumn{1}{l|}{\multirow{2}{*}{Model}}&\multicolumn{2}{c|}{Q$\rightarrow${A}}&\multicolumn{2}{c|}{QA$\rightarrow${R}}&\multicolumn{2}{c}{Q$\rightarrow${AR}}\\
                                                                                    \cmidrule(lr){3-4}          \cmidrule(lr){5-6}          \cmidrule(lr){7-8}
                                            &                                       & Val  & Test               & Val  & Test               & Val  & Test\\
    \midrule
                                            & Chance                                & 25.0 & 25.0               & 25.0 & 25.0               & 6.2  & 6.2\\
    \midrule
    \multirow{4}{0pt}{\rotatebox{90}{VQA}}  &RevisitedVQA~\cite{RevisitedVQA}       & 39.4 & 40.5               & 34.0 & 33.7               & 13.5 & 13.8\\
                                            &BUTD~\cite{BUTD}                       & 42.8 & 44.1               & 25.1 & 25.1               & 10.7 & 11.0\\
                                            &MLB~\cite{MLB}                         & 45.5 & 46.2               & 36.1 & 36.8               & 17.0 & 17.2\\
                                            &MUTAN~\cite{MUTAN}                     & 44.4 & 45.5               & 32.0 & 32.2               & 14.6 & 14.6\\
    \midrule
    \multirow{6}{0pt}{\rotatebox{90}{VCR}}  &R2C~\cite{R2C}                         & 63.8 & 65.1               & 67.2 & 67.3               & 43.1 & 44.0\\
                                            &CCN~\cite{CCN}                         & 66.6 & 67.2               & 68.3 & 68.2               & 45.5 & 46.1\\
                                            &TAB-VCR~\cite{TAB-VCR}                 & 69.9 & 70.4               & 72.2 & 71.7               & 50.6 & 50.5\\
                                            &ECMR~\cite{ECMR}                       & 70.7 & 70.4               & 72.0 & 71.3               & 51.1 & 50.5\\
                                            &VisualBERT~\cite{VisualBERT}           & 70.8 & 71.6               & 73.2 & 73.2               & 52.2 & 52.4\\
                                            &MLCC~\cite{R3_MM}                      & 71.7 & 72.3               & 73.4 & 72.8               & 52.9 & 52.8\\
                                            &VL-BERT~\cite{VL-BERT}                 & 72.6 & 73.4               & 74.0 & 74.5               & 54.0 & 54.8\\
                                            &UNITER~\cite{UNITER}                   & 74.4 & 75.5               & 76.9 & 77.3               & 57.5 & 58.6\\
    \midrule
    \multirow{3}{0pt}{\rotatebox{90}{Ours}} &$\text{R2C}_{\text{ARC}}$              & 66.2 & 66.9               & 69.1 & 69.4               & 45.9 & 46.6\\
                                            &$\text{CCN}_{\text{ARC}}$              & 69.0 & 70.0               & 71.1 & 70.6               & 49.3 & 49.7\\
                                            &$\text{TAB-VCR}_{\text{ARC}}$          & 70.5 & 71.4               & 73.1 & 72.6               & 51.8 & 52.1\\
                                            &$\text{VL-BERT}_{\text{ARC}}$          & 72.8 & -                  & 74.7 & -                  & 54.5 & -\\
                                            &$\text{UNITER}_{\text{ARC}}$           & 74.5 & -                  & 77.1 & -                  & 57.7 & -\\
    \midrule
                                            &\multicolumn{1}{l|}{Human}         & -    & 91.0      &-     & 93.0      &-     & 85.0\\
  \bottomrule
\end{tabular}}
\end{table}

The results of validation and testing sets are reported in Table 1, and the key observations are listed below.
\begin{itemize}
    \item For all five VCR baselines, with our ARC framework, they can all benefit significant gains. 
    For example, compared with CCN, 2.8$\%$, a 2.4$\%$ and 3.6$\%$ improvement on Q$\rightarrow$A, QA$\rightarrow$R, and Q$\rightarrow$AR on the test set can be observed, respectively. 
    This verifies the necessity to couple the Q$\rightarrow$A and QA$\rightarrow$R as well as the effectiveness of the proposed method. 
    \item Recent VL-Transformers, such as VL-BERT~\cite{VL-BERT} and UNITER~\cite{UNITER}, drastically outperform traditional models by large margins.
    With our ARC framework, these two models can be enhanced with further performance improvements.
    \item Traditional state-of-the-art VQA methods all perform less favorably than the VCR ones. 
    The reasons for this are two-fold: 
    1) the answers in previous VQA datasets are composed of a few keywords, while the answer length in VCR is relatively longer 
    (7.5 words for answers and 16 words for rationales on average). 
    And 2) VCR demands higher-order reasoning capability, which is far beyond the simple recognition in VQA datasets.
\end{itemize}

\subsection{Ablation Study}
We conducted detailed experiments to validate the effectiveness of each KD module and shown the results in Table~\ref{tab:ab_study}. 
From this table, we have the following observations:
\begin{table}
  \centering
  \caption{Ablation study of the proposed method on the VCR validation set. 
  \textbf{KD-R} and \textbf{KD-A} denote the KD between QR$\rightarrow${A} and QA$\rightarrow${R} and QR$\rightarrow${A} and Q$\rightarrow$A, respectively.}
  \label{tab:ab_study}
  \scalebox{1}{
  \begin{tabular}{l |c |c |c |c}
    \toprule
     Methods            & Q$\rightarrow${A}    & QA$\rightarrow${R}    & Q$\rightarrow${AR}    & QR$\rightarrow${A}\\
    \midrule
    R2C                 & 63.8                 & 67.2                  & 43.1                  & -\\
    R2C+KD-R            & 65.3                 & 68.7                  & 45.1                  & -\\
    R2C+KD-R+KD-A       & 66.2                 & 69.1                  & 45.9                  & 92.6\\
    \midrule
    CCN                 & 66.6                 & 68.3                  & 45.5                  & -\\
    CCN+KD-R            & 68.5                 & 70.7                  & 48.7                  & -\\
    CCN+KD-R+KD-A       & 69.0                 & 71.1                  & 49.3                  & 92.9\\
    \midrule
    TAB-VCR             & 69.9                 & 72.2                  & 50.6                  & -\\
    TAB-VCR+KD-R        & 70.1                 & 72.8                  & 51.3                  & -\\
    TAB-VCR+KD-R+KD-A   & 70.5                 & 73.1                  & 51.8                  & 93.1\\
  \bottomrule
\end{tabular}}
\end{table}
\begin{itemize}
    \item When incorporating the knowledge distillation module between QR$\rightarrow$A and QA$\rightarrow$R into the baseline model, a significant performance enhancement can be observed. 
    Take the R2C baseline as an example, our method boosts it by 1.5$\%$ (Q$\rightarrow$A), 1.5$\%$ (QA$\rightarrow$R) and 2.0$\%$ (Q$\rightarrow$AR). 
    \item We then introduced the knowledge distillation module between Q$\rightarrow$A and QR$\rightarrow$A to the baseline model and observed further improvement. 
    These two experiments validate the effectiveness of our knowledge distillation modules.
    \item All models perform quite well on QR$\rightarrow$A. 
    This result is intuitive since the rationale is introduced as inputs for predicting answers.
    In addition, it also supports our method implementation to make it as the proxy between Q$\rightarrow$A and QA$\rightarrow$R. 
\end{itemize}

\subsection{Qualitative Results}
\begin{figure*}[t]
  \centering
   \includegraphics[width=0.95\linewidth]{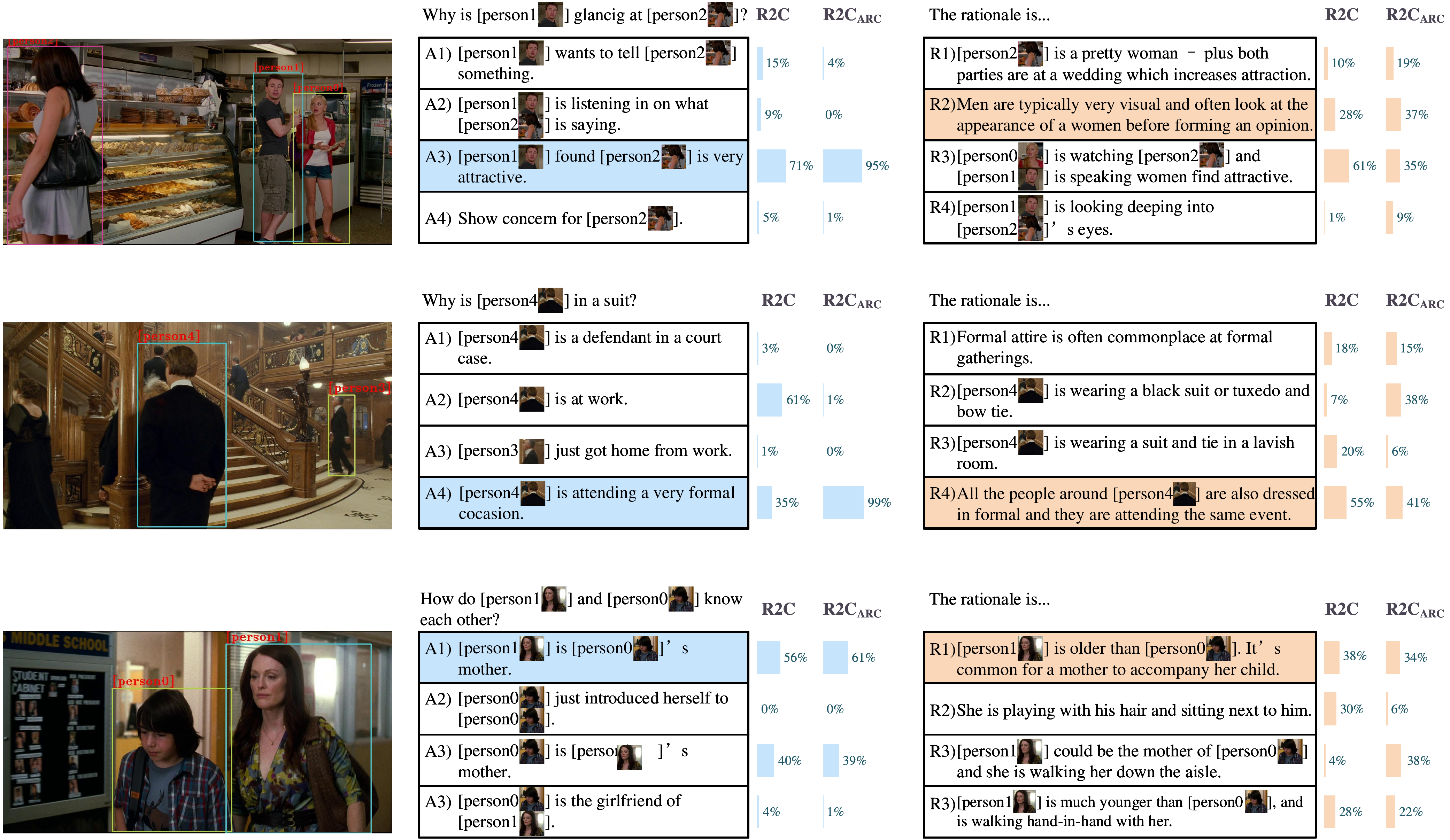}
   \caption{Qualitative results from R2C and our method. 
   The predicted probability of each option is illustrated on the rightmost.}\label{fig: case_study}
\end{figure*}

Besides the above quantitative results, we also illustrate several cases from the R2C baseline and our method in Figure~\ref{fig: case_study}. 
In detail, for the first example, although R2C gives the right answer, it fails to select the correct rationale. 
It is because the rationale option B contains more overlaps with the query (\eg `[person1]', `attractive'), and R2C leverages such a shortcut to perform reasoning. 
While with our method, the right rationale can be more confidently predicted.
The second example is even more confusing, as R2C reasons correctly, while the answer is not right. 
The key reason is that R2C takes answering and reasoning into two independent VQA instances.
The last row demonstrates a failure case, where our model chooses the right answer but predicts the wrong rationale. 
In fact, the rationale predicted by our framework is also an explanation for the right answer although it slightly deviates from the ground-truth.

\begin{figure*}[t]
  \centering
   \includegraphics[width=0.95\linewidth]{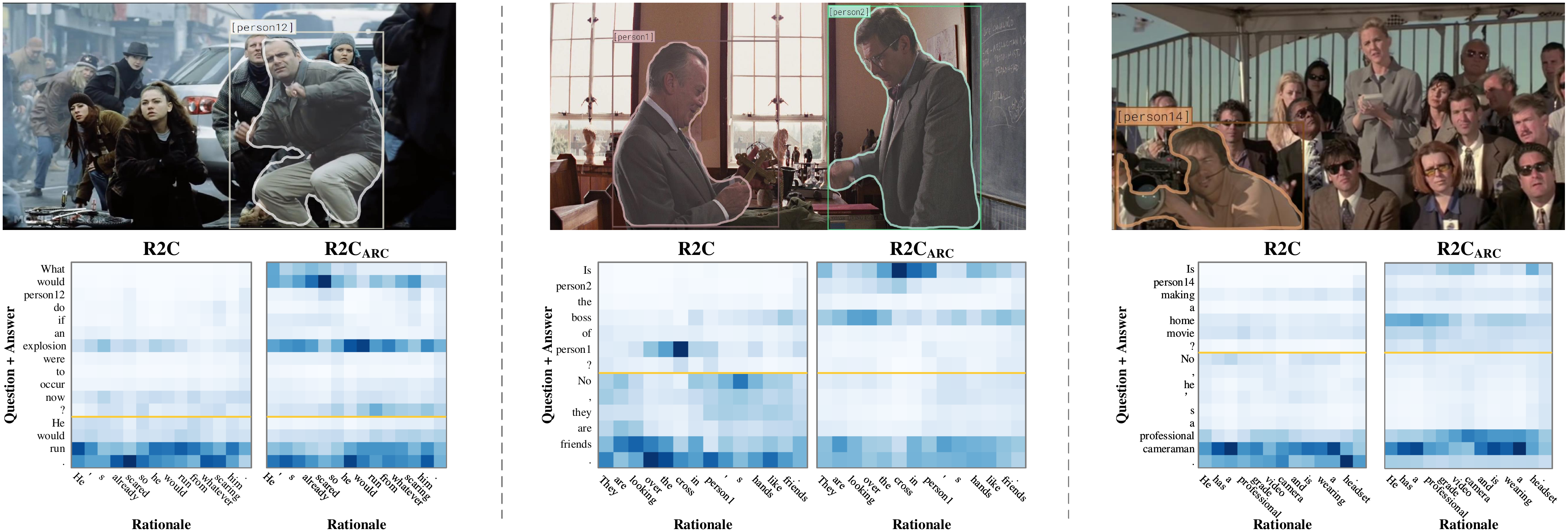}

   \caption{Attention distribution illustration from the QA$\rightarrow$R model of baselines and ours.}\label{fig: case_study_att}
\end{figure*}

Thereafter, we further illustrate two instances with attention weight distribution in Figure~\ref{fig: case_study_att}. 
From this figure, we observe that our framework guides the baseline to focus more on the question input, such as `explosion' and `boss' in the first and second example, respectively, and therefore reduces the language shortcut reliance.

\begin{table}
\centering
  \caption{Performance comparison on the out-of-domain data.}
  \label{tab:paraphrase_compare}
  \scalebox{0.85}{
  \begin{tabular}{l |c |c |c ||c |c |c}
    \toprule
    \multirow{2}*{Methods}          & \multicolumn{3}{c||}{\emph{w/o} language pre-train}                   & \multicolumn{3}{c}{\emph{w/} language pre-train}\\
                                    \cmidrule(lr){2-4}                                                      \cmidrule(lr){5-7}
                                    & Q$\rightarrow${A}    & QA$\rightarrow${R}    & Q$\rightarrow${AR}     & Q$\rightarrow${A}    & QA$\rightarrow${R}    & Q$\rightarrow${AR}\\
    \midrule
    R2C                             & 52.8                 & 54.7                  & 29.3                   & 55.2                 & 59.9                  & 33.6\\
    $\text{R2C}_{\text{ARC}}$       & 56.4                 & 57.2                  & 32.6                   & 59.3                 & 62.9                  & 37.7\\
    \midrule
    CCN                             & 52.6                 & 56.5                  & 30.4                   & 55.3                 & 60.3                  & 33.8\\
    $\text{CCN}_{\text{ARC}}$       & 55.5                 & 57.2                  & 32.6                   & 59.5                 & 65.5                  & 39.5\\
    \midrule
    TAB-VCR                         & 56.4                 & 60.5                  & 34.9                   & 60.3                 & 66.1                  & 40.4\\
    $\text{TAB-VCR}_{\text{ARC}}$   & 57.8                 & 61.3                  & 36.0                   & 62.4                 & 67.4                  & 42.7\\
  \bottomrule
\end{tabular}}
\end{table}

\subsection{Model Performance on Out-of-Domain Data}
As previously shown in Table \ref{tab:origin_vs_paraphrase}, all existing models perform much less favorably on the out-of-domain data.
To examine whether coupling answering and reasoning can improve the generalization of these models, we applied our method to these data and reported the results in Table~\ref{tab:paraphrase_compare}.
In addition, we also pre-trained the BERT model on the rewritten textual data to test whether language pre-pretraining will bring further benefits.
As can be observed, with our proposed ARC, all the three models obtain some performance improvements. 
For instance, our method gains a 3.6\%, 2.5\%, and 3.3\% absolute improvement over R2C of the \emph{w/o} language pre-train setting on the three accuracy metrics, respectively.
It is evident that our method demonstrates better generalization capability on these skewed data. 

\subsection{Results w.r.t Question Types}
\begin{figure*}[htbp]
  \centering
   \includegraphics[width=0.95\linewidth]{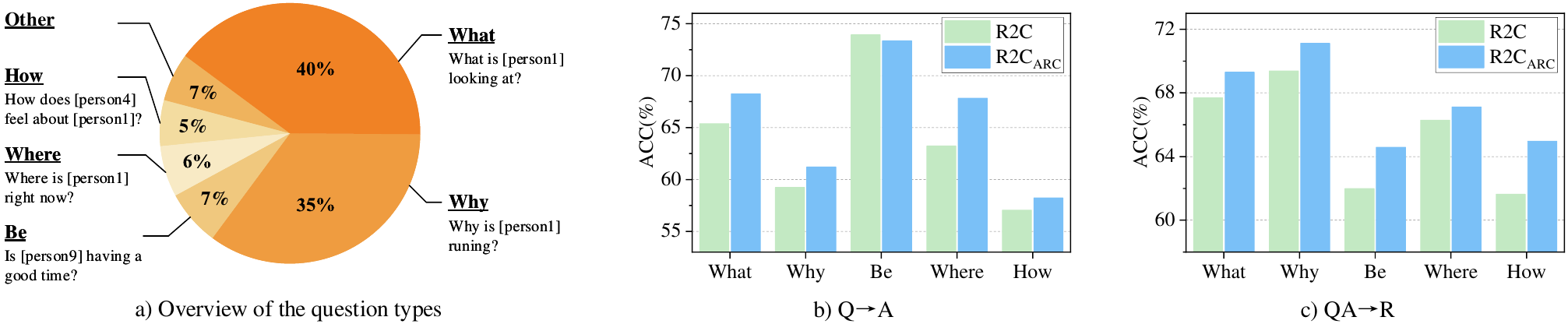}
   \caption{Validation set instance distribution and model performance according to question types.}
   \label{fig:question_type}
\end{figure*}

Figure~\ref{fig:question_type} illustrates the question types (extracted by the corresponding matching words) in the validation set.
We then show the model performance of R2C and our method with respect to question types.
In a nutshell, our method achieves consistent improvements on almost all categories.
Especially, compared with binary questions like \emph{is} and \emph{do}, our method shows more advantage on more challenging \emph{where}, \emph{how}, and \emph{what} questions.
However, both methods struggle with \emph{how} questions, as these questions demand high-level visual understanding and are therefore difficult to address.  

%% file: section/conclusion.tex
\section{Conclusion and Future Work}
\label{sec: conclusion}
Existing VCR models perform the answering and explaining processes in a separate manner, leading to poor generalization and undesirable language shortcuts between answers and rationales. 
This paper first discusses the disadvantage of the separate training strategy, followed by a novel knowledge distillation framework to couple the two processes. 
Our framework consists of two KD modules, \ie KD-A and KD-R, where the former is leveraged to align the predicted logits between Q$\rightarrow$A and QR$\rightarrow$A, 
and the latter aims to maintain semantic consistency between QA$\rightarrow$R and QR$\rightarrow$A with feature-level knowledge distillation. 
We apply this framework to several state-of-the-art baselines and studied its effectiveness on the VCR benchmark dataset. 
With the quantitative and qualitative experimental results, the viability of jointly training Q$\rightarrow$A and QR$\rightarrow$A is explicitly testified. 

Regarding the future directions, since this work demonstrates the potential of process coupling for enhancing visual understanding, studying solutions for jointly training from more model components, such as the attention module, is thus promising. 

%% file: Unified-VCR.bbl
\begin{thebibliography}{10}
\providecommand{\url}[1]{#1}
\csname url@samestyle\endcsname
\providecommand{\newblock}{\relax}
\providecommand{\bibinfo}[2]{#2}
\providecommand{\BIBentrySTDinterwordspacing}{\spaceskip=0pt\relax}
\providecommand{\BIBentryALTinterwordstretchfactor}{4}
\providecommand{\BIBentryALTinterwordspacing}{\spaceskip=\fontdimen2\font plus
\BIBentryALTinterwordstretchfactor\fontdimen3\font minus
  \fontdimen4\font\relax}
\providecommand{\BIBforeignlanguage}[2]{{%
\expandafter\ifx\csname l@#1\endcsname\relax
\typeout{** WARNING: IEEEtran.bst: No hyphenation pattern has been}%
\typeout{** loaded for the language `#1'. Using the pattern for}%
\typeout{** the default language instead.}%
\else
\language=\csname l@#1\endcsname
\fi
#2}}
\providecommand{\BIBdecl}{\relax}
\BIBdecl

\bibitem{VQA1}
S.~Antol, A.~Agrawal, J.~Lu, M.~Mitchell, D.~Batra, C.~L. Zitnick, and
  D.~Parikh, ``Vqa: Visual question answering,'' in \emph{{IEEE} International
  Conference on Computer Vision}.\hskip 1em plus 0.5em minus 0.4em\relax
  {IEEE}, 2015, pp. 2425--2433.

\bibitem{VQA2}
Y.~Zhu, O.~Groth, M.~S. Bernstein, and L.~Fei{-}Fei, ``Visual7w: Grounded
  question answering in images,'' in \emph{{IEEE} Conference on Computer Vision
  and Pattern Recognition}.\hskip 1em plus 0.5em minus 0.4em\relax {IEEE},
  2016, pp. 4995--5004.

\bibitem{R2C}
R.~Zellers, Y.~Bisk, A.~Farhadi, and Y.~Choi, ``From recognition to cognition:
  Visual commonsense reasoning,'' in \emph{{IEEE} Conference on Computer Vision
  and Pattern Recognition}.\hskip 1em plus 0.5em minus 0.4em\relax {IEEE},
  2019, pp. 6720--6731.

\bibitem{CCN}
A.~Wu, L.~Zhu, Y.~Han, and Y.~Yang, ``Connective cognition network for
  directional visual commonsense reasoning,'' in \emph{Advances in Neural
  Information Processing Systems}, 2019, pp. 5670--5680.

\bibitem{ViLBERT}
J.~Lu, D.~Batra, D.~Parikh, and S.~Lee, ``Vilbert: Pretraining task-agnostic
  visiolinguistic representations for vision-and-language tasks,'' in
  \emph{Advances in Neural Information Processing Systems}, 2019, pp. 13--23.

\bibitem{ERNIE-ViL}
F.~Yu, J.~Tang, W.~Yin, Y.~Sun, H.~Tian, H.~Wu, and H.~Wang, ``Ernie-vil:
  Knowledge enhanced vision-language representations through scene graphs,'' in
  \emph{{AAAI} Conference on Artificial Intelligence}.\hskip 1em plus 0.5em
  minus 0.4em\relax {AAAI}, 2021, pp. 3208--3216.

\bibitem{Conceptual_Captions}
P.~Sharma, N.~Ding, S.~Goodman, and R.~Soricut, ``Conceptual captions: A
  cleaned, hypernymed, image alt-text dataset for automatic image captioning,''
  in \emph{Annual Meeting of the Association for Computational
  Linguistics}.\hskip 1em plus 0.5em minus 0.4em\relax {ACL}, 2018, pp.
  2556--2565.

\bibitem{VQA3}
M.~Malinowski, M.~Rohrbach, and M.~Fritz, ``Ask your neurons: A neural-based
  approach to answering questions about images,'' in \emph{{IEEE} International
  Conference on Computer Vision}.\hskip 1em plus 0.5em minus 0.4em\relax
  {IEEE}, 2015, pp. 1--9.

\bibitem{VQA4}
Q.~Wu, D.~Teney, P.~Wang, C.~Shen, A.~R. Dick, and A.~van~den Hengel, ``Visual
  question answering: A survey of methods and datasets,'' \emph{Computer Vision
  and Image Understanding}, vol. 163, pp. 21--40, 2017.

\bibitem{BUTD}
P.~Anderson, X.~He, C.~Buehler, D.~Teney, M.~Johnson, S.~Gould, and L.~Zhang,
  ``Bottom-up and top-down attention for image captioning and visual question
  answering,'' in \emph{{IEEE} Conference on Computer Vision and Pattern
  Recognition}.\hskip 1em plus 0.5em minus 0.4em\relax {IEEE}, 2018, pp.
  6077--6086.

\bibitem{Re-attention}
W.~Guo, Y.~Zhang, J.~Yang, and X.~Yuan, ``Re-attention for visual question
  answering,'' \emph{IEEE Transactions on Image Processing}, vol.~30, pp.
  6730--6743, 2021.

\bibitem{NMN}
J.~Andreas, M.~Rohrbach, T.~Darrell, and D.~Klein, ``Neural module networks,''
  in \emph{{IEEE} Conference on Computer Vision and Pattern Recognition}.\hskip
  1em plus 0.5em minus 0.4em\relax {IEEE}, 2016, pp. 39--48.

\bibitem{Clevr}
J.~Johnson, B.~Hariharan, L.~van~der Maaten, L.~Fei{-}Fei, C.~L. Zitnick, and
  R.~B. Girshick, ``Clevr: A diagnostic dataset for compositional language and
  elementary visual reasoning,'' in \emph{{IEEE} Conference on Computer Vision
  and Pattern Recognition}.\hskip 1em plus 0.5em minus 0.4em\relax {IEEE},
  2017, pp. 1988--1997.

\bibitem{Fvqa}
P.~Wang, Q.~Wu, C.~Shen, A.~R. Dick, and A.~van~den Hengel, ``Fvqa: Fact-based
  visual question answering,'' \emph{IEEE Transactions on Pattern Analysis and
  Machine Intelligence}, vol.~40, no.~10, pp. 2413--2427, 2018.

\bibitem{Ok-vqa}
K.~Marino, M.~Rastegari, A.~Farhadi, and R.~Mottaghi, ``Ok-vqa: A visual
  question answering benchmark requiring external knowledge,'' in \emph{{IEEE}
  Conference on Computer Vision and Pattern Recognition}.\hskip 1em plus 0.5em
  minus 0.4em\relax {IEEE}, 2019, pp. 3195--3204.

\bibitem{HINT}
R.~R. Selvaraju, S.~Lee, Y.~Shen, H.~Jin, S.~Ghosh, L.~P. Heck, D.~Batra, and
  D.~Parikh, ``Taking a hint: Leveraging explanations to make vision and
  language models more grounded,'' in \emph{{IEEE} International Conference on
  Computer Vision}.\hskip 1em plus 0.5em minus 0.4em\relax {IEEE}, 2019, pp.
  2591--2600.

\bibitem{Adavqa}
Y.~Guo, L.~Nie, Z.~Cheng, F.~Ji, J.~Zhang, and A.~D. Bimbo, ``Adavqa:
  Overcoming language priors with adapted margin cosine loss,'' in
  \emph{International Joint Conference on Artificial Intelligence}.\hskip 1em
  plus 0.5em minus 0.4em\relax Morgan Kaufmann, 2021, pp. 708--714.

\bibitem{Dont_just_assume}
A.~Agrawal, D.~Batra, D.~Parikh, and A.~Kembhavi, ``Don't just assume; look and
  answer: Overcoming priors for visual question answering,'' in \emph{{IEEE}
  Conference on Computer Vision and Pattern Recognition}.\hskip 1em plus 0.5em
  minus 0.4em\relax {IEEE}, 2018, pp. 4971--4980.

\bibitem{VQA-E}
Q.~Li, Q.~Tao, S.~R. Joty, J.~Cai, and J.~Luo, ``Vqa-e: Explaining,
  elaborating, and enhancing your answers for visual questions,'' in
  \emph{European Conference on Computer Vision}.\hskip 1em plus 0.5em minus
  0.4em\relax Springer, 2018, pp. 570--586.

\bibitem{Explanation_vs_attention}
B.~N. Patro, Anupriy, and V.~P. Namboodiri, ``Explanation vs attention: A
  two-player game to obtain attention for vqa,'' in \emph{{AAAI} Conference on
  Artificial Intelligence}.\hskip 1em plus 0.5em minus 0.4em\relax {AAAI},
  2020, pp. 11\,848--11\,855.

\bibitem{Grad-cam}
R.~R. Selvaraju, M.~Cogswell, A.~Das, R.~Vedantam, D.~Parikh, and D.~Batra,
  ``Grad-cam: Visual explanations from deep networks via gradient-based
  localization,'' in \emph{{IEEE} International Conference on Computer
  Vision}.\hskip 1em plus 0.5em minus 0.4em\relax {IEEE}, 2017, pp. 618--626.

\bibitem{U-cam}
B.~N. Patro, M.~Lunayach, S.~Patel, and V.~P. Namboodiri, ``U-cam: Visual
  explanation using uncertainty based class activation maps,'' in \emph{{IEEE}
  International Conference on Computer Vision}.\hskip 1em plus 0.5em minus
  0.4em\relax {IEEE}, 2019, pp. 7443--7452.

\bibitem{VQA-HAT}
A.~Das, H.~Agrawal, L.~Zitnick, D.~Parikh, and D.~Batra, ``Human attention in
  visual question answering: Do humans and deep networks look at the same
  regions?'' \emph{Computer Vision and Image Understanding}, vol. 163, pp.
  90--100, 2017.

\bibitem{VQA-X}
D.~H. Park, L.~A. Hendricks, Z.~Akata, A.~Rohrbach, B.~Schiele, T.~Darrell, and
  M.~Rohrbach, ``Multimodal explanations: Justifying decisions and pointing to
  the evidence,'' in \emph{{IEEE} Conference on Computer Vision and Pattern
  Recognition}.\hskip 1em plus 0.5em minus 0.4em\relax {IEEE}, 2018, pp.
  8779--8788.

\bibitem{VQA5}
J.~Wu and R.~J. Mooney, ``Faithful multimodal explanation for visual question
  answering,'' in \emph{Workshop on Annual Meeting of the Association for
  Computational Linguistics}.\hskip 1em plus 0.5em minus 0.4em\relax {ACL},
  2019, pp. 103--112.

\bibitem{HGL}
W.~Yu, J.~Zhou, W.~Yu, X.~Liang, and N.~Xiao, ``Heterogeneous graph learning
  for visual commonsense reasoning,'' in \emph{Advances in Neural Information
  Processing Systems}, 2019, pp. 2765--2775.

\bibitem{ECMR}
X.~Zhang, F.~Zhang, and C.~Xu, ``Explicit cross-modal representation learning
  for visual commonsense reasoning,'' \emph{IEEE Transactions on Multimedia},
  vol.~24, pp. 2986--2997, 2022.

\bibitem{R3_5_ref2}
J.~Zhu and H.~Wang, ``Multiscale conditional relationship graph network for
  referring relationships in images,'' \emph{{IEEE} Transactions on Cognitive
  and Development Systems}, vol.~14, no.~2, pp. 752--760, 2022.

\bibitem{R3_MM}
X.~Zhang, F.~Zhang, and C.~Xu, ``Multi-level counterfactual contrast for visual
  commonsense reasoning,'' in \emph{International Conference on
  Multimedia}.\hskip 1em plus 0.5em minus 0.4em\relax {ACM}, 2021, pp.
  1793--1802.

\bibitem{UNITER}
Y.~Chen, L.~Li, L.~Yu, A.~E. Kholy, F.~Ahmed, Z.~Gan, Y.~Cheng, and J.~Liu,
  ``Uniter: Universal image-text representation learning,'' in \emph{European
  Conference on Computer Vision}.\hskip 1em plus 0.5em minus 0.4em\relax
  Springer, 2020, pp. 104--120.

\bibitem{VL-BERT}
W.~Su, X.~Zhu, Y.~Cao, B.~Li, L.~Lu, F.~Wei, and J.~Dai, ``Vl-bert:
  Pre-training of generic visual-linguistic representations,'' in
  \emph{International Conference on Learning Representations}, 2020.

\bibitem{12-in-1}
J.~Lu, V.~Goswami, M.~Rohrbach, D.~Parikh, and S.~Lee, ``12-in-1: Multi-task
  vision and language representation learning,'' in \emph{Computer Vision and
  Pattern Recognition}.\hskip 1em plus 0.5em minus 0.4em\relax {IEEE}, 2020,
  pp. 10\,434--10\,443.

\bibitem{MERLOT_RESERVE}
R.~Zellers, J.~Lu, X.~Lu, Y.~Yu, Y.~Zhao, M.~Salehi, A.~Kusupati, J.~Hessel,
  A.~Farhadi, and Y.~Choi, ``Merlot reserve: Neural script knowledge through
  vision and language and sound,'' in \emph{{IEEE} Conference on Computer
  Vision and Pattern Recognition}.\hskip 1em plus 0.5em minus 0.4em\relax
  {IEEE}, 2022, pp. 16\,354--16\,366.

\bibitem{KD}
G.~E. Hinton, O.~Vinyals, and J.~Dean, ``Distilling the knowledge in a neural
  network,'' \emph{CoRR}, vol. abs/1503.02531, 2015.

\bibitem{Lipschitz}
Y.~Shang, B.~Duan, Z.~Zong, L.~Nie, and Y.~Yan, ``Lipschitz continuity guided
  knowledge distillation,'' in \emph{{IEEE} International Conference on
  Computer Vision}.\hskip 1em plus 0.5em minus 0.4em\relax {IEEE}, 2021, pp.
  10\,675--10\,684.

\bibitem{KD_Logits}
J.~Ba and R.~Caruana, ``Do deep nets really need to be deep?'' in
  \emph{Advances in Neural Information Processing Systems}, 2014, pp.
  2654--2662.

\bibitem{KTAN}
P.~Liu, W.~Liu, H.~Ma, Z.~Jiang, and M.~Seok, ``Ktan: Knowledge transfer
  adversarial network,'' in \emph{{IEEE} International Joint Conference on
  Neural Networks}.\hskip 1em plus 0.5em minus 0.4em\relax {IEEE}, 2020, pp.
  1--7.

\bibitem{MEAL}
Z.~Shen, Z.~He, and X.~Xue, ``Meal: Multi-model ensemble via adversarial
  learning,'' in \emph{{AAAI} Conference on Artificial Intelligence}.\hskip 1em
  plus 0.5em minus 0.4em\relax {AAAI}, 2019, pp. 4886--4893.

\bibitem{FitNets}
A.~Romero, N.~Ballas, S.~E. Kahou, A.~Chassang, C.~Gatta, and Y.~Bengio,
  ``Fitnets: Hints for thin deep nets,'' in \emph{International Conference on
  Learning Representations}, 2015.

\bibitem{Paying_More_Attention_to_Attention}
S.~Zagoruyko and N.~Komodakis, ``Paying more attention to attention: Improving
  the performance of convolutional neural networks via attention transfer,'' in
  \emph{International Conference on Learning Representations}, 2017.

\bibitem{RCO}
X.~Jin, B.~Peng, Y.~Wu, Y.~Liu, J.~Liu, D.~Liang, J.~Yan, and X.~Hu,
  ``Knowledge distillation via route constrained optimization,'' in
  \emph{{IEEE} Conference on Computer Vision and Pattern Recognition}.\hskip
  1em plus 0.5em minus 0.4em\relax {IEEE}, 2019, pp. 1345--1354.

\bibitem{FSP}
J.~Yim, D.~Joo, J.~Bae, and J.~Kim, ``A gift from knowledge distillation: Fast
  optimization, network minimization and transfer learning,'' in \emph{{IEEE}
  Conference on Computer Vision and Pattern Recognition}.\hskip 1em plus 0.5em
  minus 0.4em\relax {IEEE}, 2017, pp. 7130--7138.

\bibitem{rescale-prior}
Y.~Guo, L.~Nie, Z.~Cheng, Q.~Tian, and M.~Zhang, ``Loss re-scaling {VQA:}
  revisiting the language prior problem from a class-imbalance view,''
  \emph{IEEE Transactions on Image Processing}, vol.~31, pp. 227--238, 2022.

\bibitem{TAB-VCR}
J.~Lin, U.~Jain, and A.~G. Schwing, ``Tab-vcr: Tags and attributes based vcr
  baselines,'' in \emph{Advances in Neural Information Processing Systems},
  2019, pp. 15\,589--15\,602.

\bibitem{LSMDC}
A.~Rohrbach, A.~Torabi, M.~Rohrbach, N.~Tandon, C.~J. Pal, H.~Larochelle, A.~C.
  Courville, and B.~Schiele, ``Movie description,'' \emph{International Journal
  of Computer Vision}, vol. 123, no.~1, pp. 94--120, 2017.

\bibitem{MASK_RCNN}
K.~He, G.~Gkioxari, P.~Doll{\'{a}}r, and R.~B. Girshick, ``Mask r-cnn,'' in
  \emph{{IEEE} International Conference on Computer Vision}.\hskip 1em plus
  0.5em minus 0.4em\relax {IEEE}, 2017, pp. 2980--2988.

\bibitem{RevisitedVQA}
A.~Jabri, A.~Joulin, and L.~van~der Maaten, ``Revisiting visual question
  answering baselines,'' in \emph{European Conference on Computer
  Vision}.\hskip 1em plus 0.5em minus 0.4em\relax Springer, 2016, pp. 727--739.

\bibitem{MLB}
J.~Kim, K.~W. On, W.~Lim, J.~Kim, J.~Ha, and B.~Zhang, ``Hadamard product for
  low-rank bilinear pooling,'' in \emph{International Conference on Learning
  Representations}, 2017.

\bibitem{MUTAN}
H.~Ben{-}younes, R.~Cad{\`{e}}ne, M.~Cord, and N.~Thome, ``Mutan: Multimodal
  tucker fusion for visual question answering,'' in \emph{{IEEE} International
  Conference on Computer Vision}.\hskip 1em plus 0.5em minus 0.4em\relax
  {IEEE}, 2017, pp. 2631--2639.

\bibitem{VisualBERT}
L.~H. Li, M.~Yatskar, D.~Yin, C.~Hsieh, and K.~Chang, ``Visualbert: A simple
  and performant baseline for vision and language,'' \emph{CoRR}, vol.
  abs/1908.03557, 2019.

\end{thebibliography}
